# Mid-Long Term Daily Electricity Consumption Forecasting Based on Piecewise Linear Regression and Dilated Causal CNN

LAN Zhou[1], LIU Ben[2], FENG Yi[3], DONG Danhuang[1], ZHANG Peng[2]

**Abstract:** Daily electricity consumption forecasting is a classical problem. Existing forecasting algorithms tend to have decreased accuracy on special dates like holidays. This study decomposes the daily electricity consumption series into three components: trend, seasonal, and residual, and constructs a two-stage prediction method using piecewise linear regression as a filter and Dilated Causal CNN as a predictor. The specific steps involve setting breakpoints on the time axis and fitting the piecewise linear regression model with one-hot encoded information such as month, weekday, and holidays. For the challenging prediction of the Spring Festival, distance is introduced as a variable using a third-degree polynomial form in the model. The residual sequence obtained in the previous step is modeled using Dilated Causal CNN, and the final prediction of daily electricity consumption is the sum of the two-stage predictions. Experimental results demonstrate that this method achieves higher accuracy compared to existing approaches.

**Key words:** Daily electricity consumption forecasting; time series decomposition; piecewise linear regression; Dilated Causal CNN.

## 1 Introduction

The prediction of daily electricity consumption plays a crucial guiding role in the production, transportation, and pricing of the electricity industry, making it one of the classic problems in the field. There are roughly three categories of time series forecasting methods: based on classical stationary time series theory; based on statistical learning and machine learning methods; and based on deep learning frameworks. In general, directly applying the above-mentioned methods to daily electricity consumption sequence may not yield satisfactory results. Therefore, previous scholars have conducted the following researches: incorporating factors that have a significant impact on daily electricity consumption into the model, such as economic factors[1] and weather conditions[2]; decomposing the daily electricity consumption sequence and separately forecasting each component before summing them up, such as X-12-ARIMA[3], seasonal-trend decomposition[4], empirical mode decomposition[5], and variational mode decomposition[6]; combining multiple models[7-11].

In summary, the prediction of daily electricity consumption is generally a medium to long-term forecast, such as forecasting the daily consumption for the next few months. The combination of medium to long-term forecasting methods with deep learning frameworks is not as closely integrated as short-term forecasting. Instead, it often involves decomposing the sequence into several sub-sequences and then applying traditional models suitable for short-term forecasting to predict these sub-sequences. The improvement in prediction accuracy primarily depends on the introduction or adjustment of additional variables. This approach fails to capture irregular low-density information in long-term time series forecasting, such as the Chinese New Year. To address this issue, this paper proposes a two-stage prediction method that separately adjusts for holiday and similar information to achieve better forecasting results.

## 2 Methods and Procedures

### 2.1 Two-Stage Idea

Considering $X_t$ as the daily electricity consumption sequence, it is subjected to the following classical time series decomposition[12]:

$$X_t = T_t(d) + P_t(s) + \varepsilon_t \quad (1)$$

$T_t(d)$ represents a $d$-order polynomial sequence, $P_t(s)$ represents a periodically spaced sequence with a period of $s$, and $\varepsilon_t$ represents the residual sequence. At this point, the trend and periodic components of the daily electricity consumption sequence can be eliminated through differencing operations:

$$\nabla^d \nabla_s X_t = \varepsilon_t \quad (2)$$

$\nabla$ represents the difference operator, $\nabla^d$ indicates taking a $d$-order difference, and $\nabla_s$ indicates taking a difference with a span of $s$.

After removing the trend and periodic components from the sequence, if at this point all information has been extracted, $\varepsilon_t$ represents a noise sequence and no further operation is required. However, if $\varepsilon_t$ follows a stationary process, then classical stationary time series theory can be used for parameter estimation and forecasting. This is the essence of the two-stage forecasting approach.

## 2.2 Filter Based on Piecewise Linear Regression

For actual daily electricity consumption sequences, $T_t(d)$ and $P_t(s)$ often do not strictly adhere to classical assumptions, resulting in a different decomposition of $X_t$:

$$X_t = T_t + P_t + \varepsilon_t \quad (3)$$

$T_t$ represents a complex time trend sequence, while $P_t$ represents a pseudo-periodic sequence, meaning it exhibits an approximate periodicity but the periods are not equidistant. In this case, it is not possible to fit $T_t$ and $P_t$ directly through differencing operations, and alternative methods need to be sought.

### 2.2.1 Tread Fitting

For the trend, different modes may appear during different time periods within a year. The daily electricity consumption for a certain region from 2018 to 2021 is shown in Figure 1.

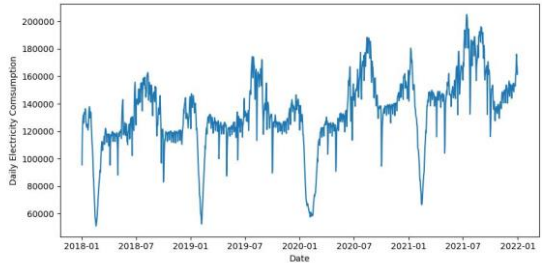

Fig. 1 Daily electricity consumption series from 2018 to 2021

From Figure 1, it can be observed that, if we exclude the impact of the Chinese New Year from January to March, we can roughly divide the daily electricity consumption sequence for each year into six segments: the first segment is from January to March at the beginning of the year, continuing from the previous year, and the data shows a slight downward trend; the second segment is from March to June, during the spring season when flowers bloom, and the data remains relatively stable or shows a slight upward trend; the third segment is from June to August, characterized by a heatwave, where the data rapidly increases; the fourth segment is from August to October when temperatures start to fall, and the data decreases rapidly; the fifth segment is from October to November, with clear autumn weather, and the data remains stable; the sixth segment is from November to December, as the year-end period, and the data shows a slight increase.

In this case, if a single polynomial function is used to fit such a time trend, it would require a high degree $d$, increasing the complexity of the model and reducing its generalization ability. Instead, considering it as a piecewise linear trend appears to be more suitable. Therefore, this paper selects piecewise linear regression for daily electricity consumption filtering. Assuming $p$ breakpoints are set for $X_t$, with date $t$ as the explanatory variable, the resulting piecewise regression equation is:

$$X_t = \beta_0 + \beta_1 t + \sum_{i=1}^{p} \beta_{i+1}(t - t_i)I(t > t_i) \quad (4)$$

$t_i$ represents the breakpoints, and $I(\cdot)$ is the indicator function.

### 2.2.2 Periodic Correction

The characteristics that affect the daily electricity consumption sequence in terms of periodic influences are mainly the month, the day of the week, and holidays. Among these, months and days of the week follow equidistant periods, while some holidays exhibit pseudo-periods. To account for these periodic components as well as adjustment days, it is considered to introduce them as dummy variables using one-hot encoding. Here, adjustment days refer to those that would normally be weekends but become working days due to holidays. In this case, the regression equation is adjusted to:

$$X_t = \beta_0 + \beta_1 t + \sum_{i=1}^{p} \beta_{i+1}(t - t_i)I(t > t_i) \\ + f(M_t) + f(W_t) + f(H_t) + f(C_t) \quad (5)$$

$M_t$ represents the month variable, $W_t$ represents the day of the week variable, $H_t$ represents the holiday variable, $C_t$ represents the adjustment day variable, and $f(\cdot)$ is the function used to one-hot encode the variables.

### 2.2.3 Spring Festival Correction

As the most significant traditional festival for the Chinese people, the Spring Festival holds extraordinary significance and is a critical factor to focus on when forecasting electricity load. From Figure 1, it can be observed that during the Spring Festival period, there is a significant drop in the daily electricity consumption curve. Its impact extends beyond the official seven-day holiday and affects a period both before and after. After further experimental analysis, it was found that, centered around the 4th day of the Spring Festival holiday, the influence lasts for approximately 21 days in total. Therefore, unlike other official holidays, the Spring Festival needs to be considered as an independent variable.

From Figure 1, it can also be seen that the Spring Festival holiday has varying degrees of impact during its early, middle, and late stages, with the midpoint having the most significant effect and the influence gradually diminishing as you move further away from the middle of the holiday. This property cannot be captured by dummy variables because introducing dummy variables assumes that the impact is the same for each day during the holiday. Therefore, taking the 4th day as the center, this paper introduces the Spring Festival holiday as a continuous variable $S$, which represents the distance to the center of the Spring Festival holiday:

$$S_t = \begin{cases} |t - S_c| & |t - S_c| \leq 21 \\ 0 & |t - S_c| > 21 \end{cases} \quad (6)$$

$S_c$ represents the center of the Spring Festival. Then, the variable $S$ is introduced into the regression model as a third-order polynomial, and the final regression equation is adjusted to:

$$\begin{aligned} X_t = \beta_0 + \beta_1 t + \sum_{i=1}^{p} \beta_{i+1}(t - t_i)I(t > t_i) \\ + f(M_t) + f(W_t) + f(H'_t) + f(C_t) \\ + \alpha_0 S_t^3 + \alpha_1 S_t^2 + \alpha_2 S_t \end{aligned} \quad (7)$$

$H'_t$ represents holiday variables other than the Spring Festival.

### 2.3 Residual Prediction Base on Dilated Causal CNN

Through the piecewise linear regression, the daily electricity consumption sequence has largely eliminated the trend and periodic components, resulting in a residual sequence $\varepsilon_t$. However, in general, the residual sequence $\varepsilon_t$ at this stage is not entirely random and still contains learnable information. The assumption of stationarity is often too strong, and $\varepsilon_t$ tends to be more complex. Therefore, we consider further analysis based on a deep learning framework and proceed with the following modeling:

$$\varepsilon_t = f(\varepsilon_{t-1}, \varepsilon_{t-2}, \cdots, \varepsilon_{t-L}) + e_t \quad (8)$$

$e_t$ represents the noise sequence, and $f(\cdot)$ is a nonlinear function.

Considering that the daily electricity consumption of the current day is related to the past week, the past month, and even the previous year, there may still be such relationships hidden within the residual sequence. Therefore, this paper employs Dilated Causal CNN[13] to model the residual sequence. The model architecture includes 8 Dilated Causal CNN layers and two Dense layers. The results are output through a Lambda layer, and the order of these layers is shown in Figure 2. The model uses mean squared error as the loss function, utilizes the Adam optimizer, and applies a dropout rate of 0.2 to prevent overfitting.

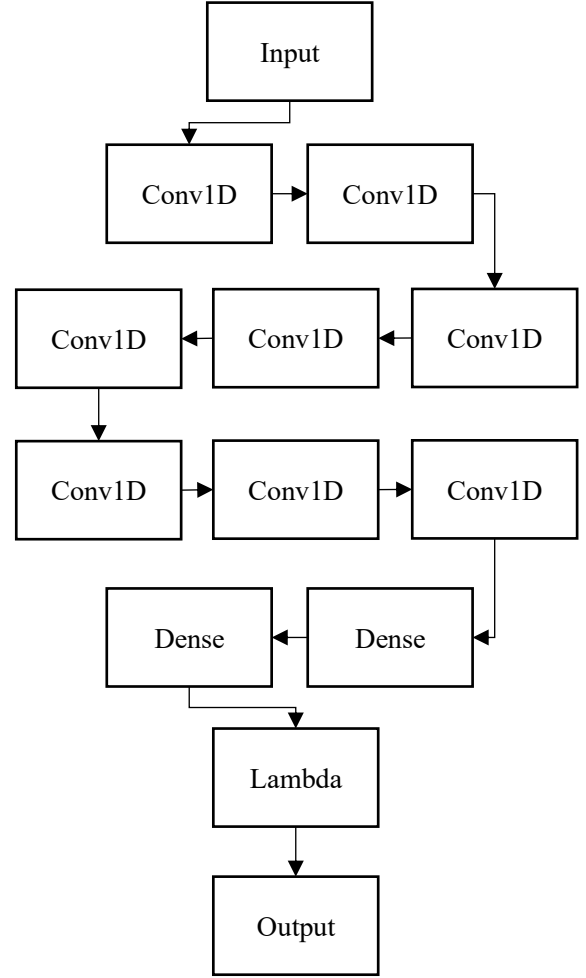

Fig. 2 Structure used in this paper

### 2.4 Daily Electricity Consumption Prediction

Combining the trend and periodic fitting and the residual prediction, the final daily electricity consumption forecast is the sum of the predictions from both methods, as follows:

$$X_t = X_{pwl,t} + \varepsilon_{cnn,t} \quad (9)$$

$X_{pwl,t}$ represents the piecewise linear regression prediction at time $t$, and $\varepsilon_{cnn,t}$ represents the residual prediction at time $t$. It's worth noting that when using the fitted piecewise linear regression model for forecasting, it's essential to first determine if there are breakpoints within the forecast horizon. If breakpoints exist, further analysis is required to estimate the location of the breakpoint and the slope of the curve after the breakpoint. The final model framework is illustrated in Figure 3.

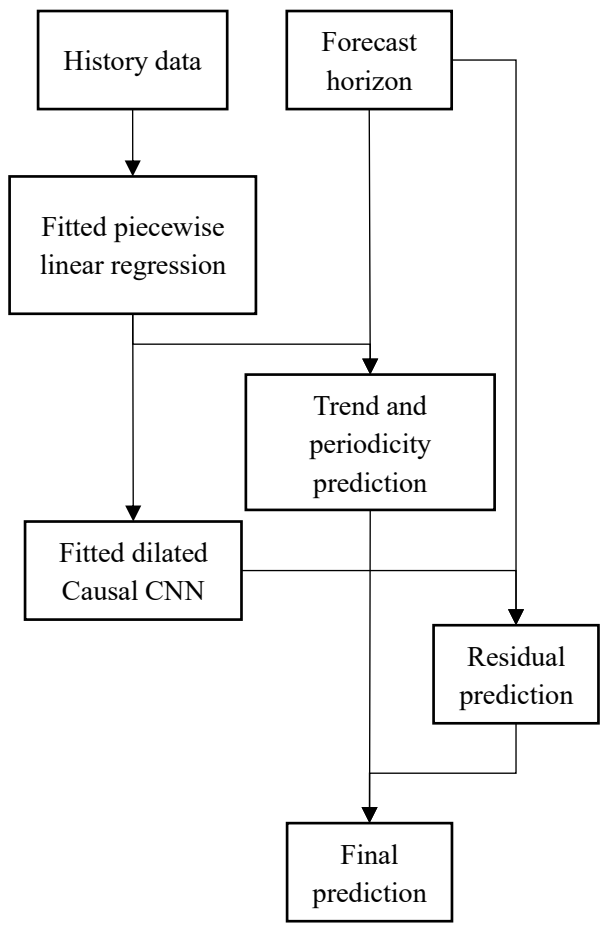

Fig. 3 Structure of prediction model used in this paper

## 3 Results

This study collected daily electricity consumption data from January 2018 to March 2022 for a certain region. The sequences from 2018 to 2021 were used as the training dataset to build the model, while the first three months of 2022 were used as the test dataset to evaluate the model, as shown in Figure 4. It's important to note that the data from January to March 2020 was excluded due to the impact of the COVID-19 pandemic.

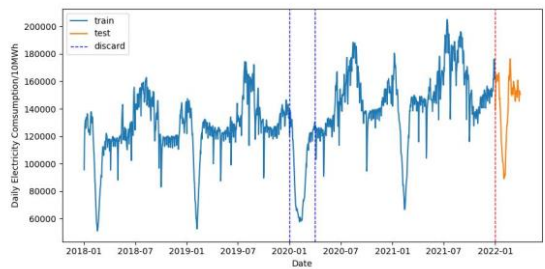

Fig. 4 Train and test set of daily electricity consumption

### 3.1 Filter

Before performing piecewise linear regression on the daily electricity consumption, it's essential to determine the number and positions of breakpoints. The selection of the quantity and location of $t_i$ can be guided by the literature [14], or it can be determined by professionals based on their practical experience. In this paper, breakpoints were manually assigned, and the positions of breakpoints for the training dataset are shown in Figure 5.

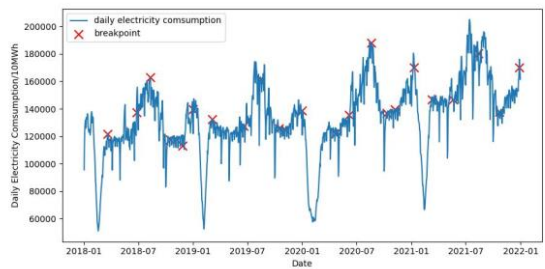

Fig. 5 Breakpoint of train set

The fitting results of piecewise linear regression and the residual sequence are shown in Figure 6.

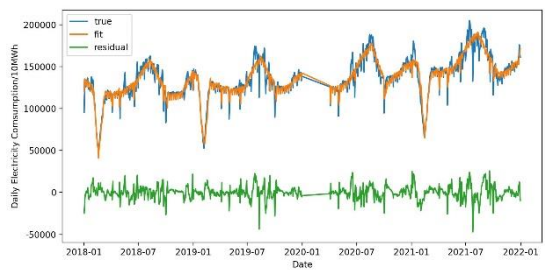

Fig. 6 Fit and residual of piecewise linear regression

From Figure 6, it can be seen that the residual sequence no longer exhibits obvious trend and periodic components but oscillates around zero. When making predictions, based on the distribution of breakpoints over the past four years, it is estimated that there will be two breakpoints within the next three months. Therefore, January 1, 2022, and March 10, 2022, were set as breakpoints. The trajectory of the curve after the breakpoints is expected to be similar to the same period in the past four years. Based on the estimated curve slopes of the past four years, the slope after the first breakpoint is approximately -300, and after the second breakpoint, it is approximately 200. The final prediction of piecewise linear regression are shown in Figure 7.

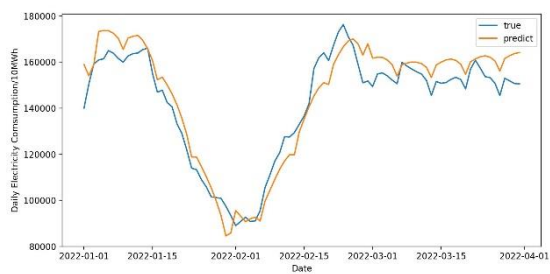

Fig. 7 Prediction of piecewise linear regression

From Figure 7, it can be observed that there is a small difference between the predicted values and the actual values. Piecewise linear regression effectively captures the trend and periodic information in the daily electricity consumption sequence, serving as a filtering mechanism.

### 3.2 Daily Electricity Consumption Prediction

Combining residual prediction, the daily electricity consumption forecasting curve on the test dataset is shown in Figure 8.

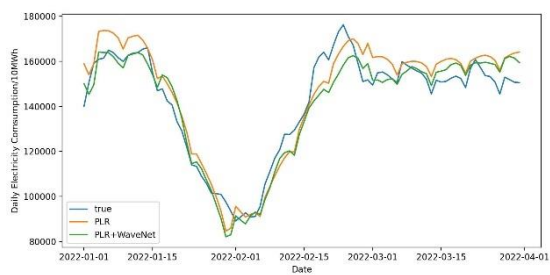

Fig. 8 Predicted curve of daily power

From Figure 8, it can be observed that compared to the prediction of piecewise linear regression alone, the residual prediction serves as a refinement, making the predicted curve closer to the actual curve. The error comparison between different prediction models is shown in Table 1, and the model parameters were obtained through multiple training and optimization adjustments.

Table 1 Comparison of different models

| Model | RMSE | MAPE/% |
|---|---|---|
| LSTM | 19115 | 10.96 |
| Dilated Causal CNN | 15491 | 9.28 |
| PLR | 7915 | 4.94 |
| PLR+ARIMA | 7567 | 4.80 |
| PLR+LSTM | 7473 | 4.69 |
| PLR+ Dilated Causal CNN | 7220 | 3.99 |

Table 1 shows that piecewise linear regression significantly improves the prediction accuracy. This improvement is attributed to the modeling approach that combines human prior knowledge, which can handle more complex patterns compared to directly applying deep learning to the original time series. Among all the prediction methods, the combination of piecewise linear regression and Dilated Causal CNN produces the best results.

## 4 Conclusion

For the daily electricity consumption sequence, this paper decomposes it into three parts: the trend component $T_t$, the periodic component $P_t$, and the residual component $\varepsilon_t$. Piecewise linear regression is used to fit the trend and periodic components, while Dilated Causal CNN is employed to model the residuals obtained from regression. The final prediction of daily electricity consumption is the sum of these two prediction parts. In terms of details, for the challenging prediction of the Spring Festival, distance is used as a variable and introduced into the regression model in the form of a third-degree polynomial. Empirical analysis shows that this method significantly improves prediction accuracy compared to existing methods and has practical value.